\documentclass[10pt,twocolumn,letterpaper]{article}

\usepackage{iccv}
\usepackage{times}
\usepackage{epsfig}
\usepackage{graphicx}
\usepackage{amsmath}
\usepackage{amssymb}
\usepackage{multicol}
\usepackage{multirow}
\usepackage{bigstrut}
\usepackage[linesnumbered,boxed,ruled,commentsnumbered]{algorithm2e}

\usepackage[pagebackref=true,breaklinks=true,letterpaper=true,colorlinks,bookmarks=false]{hyperref}

\iccvfinalcopy 

\def\iccvPaperID{9753} 

\ificcvfinal\fi

\newcommand{\rz}[1]{\textcolor{black}{#1}}

\begin{document}

\title{Rehearsal-Free Domain Continual Face Anti-Spoofing: Generalize More and Forget Less}

\author{Rizhao Cai\\
{\tt\small rzcai@ntu.edu.sg}
\and
Yawen Cui\\
{\tt\small yawen.cui@oulu.fi}
\and
Zhi Li\\
{\tt\small zhi.li@e.ntu.edu.sg}
\and
Zitong Yu$^*$\\
{\tt\small yuzitong@gbu.edu.cn}
\and
Haoliang Li\\
{\tt\small haoliang.li@cityu.edu.hk}
\and
Yongjian Hu\\
{\tt\small eeyjhu@scut.edu.cn}
\and
Alex Kot\\
{\tt\small eackot@ntu.edu.sg}
}

\maketitle
\ificcvfinal\thispagestyle{empty}\fi

\vspace{-1em}
\begin{abstract}
\vspace{-1em}
Face Anti-Spoofing (FAS) is recently studied under the continual learning setting, where the FAS models are expected to evolve after encountering the data from new domains. However, existing methods need extra replay buffers to store previous data for rehearsal, which becomes infeasible when previous data is unavailable because of privacy issues. In this paper, we propose the first rehearsal-free method for Domain Continual Learning (DCL) of FAS, which deals with catastrophic forgetting and unseen domain generalization problems simultaneously. For better generalization to unseen domains, we design the Dynamic Central Difference Convolutional Adapter (DCDCA) to adapt Vision Transformer (ViT) models during the continual learning sessions. To alleviate the forgetting of previous domains without using previous data, we propose the Proxy Prototype Contrastive Regularization (PPCR) to constrain the continual learning with previous domain knowledge from the proxy prototypes. Simulate practical DCL scenarios, we devise two new protocols which evaluate both generalization and anti-forgetting performance. Extensive experimental results show that our proposed method can improve the generalization performance in unseen domains and alleviate the catastrophic forgetting of the previous knowledge. The codes and protocols will be released soon.
\end{abstract}
\vspace{-2.5em}

\section{Introduction}\label{sec:intro}
\vspace{-0.3em}
Face recognition (FR) has been widely used in identity authentication because of its convenience. However, face recognition-based authentication systems are threatened by face spoofing attacks \cite{ramachandra2017presentation, kong2022digital, yu2022deep}. To protect FR systems from spoofing attacks, Face Anti-Spoofing (FAS) techniques are deployed to detect spoofing faces and reject malicious attempts. Although recent FAS methods based on deep learning and neural networks achieve exquisite accuracy in intra-domain testing, the performance of existing methods heavily relies on the diversity of the training data and degrades severely if there are domain shifts between the training and testing data domains \cite{li2018unsupervised}. The cross-domain problem hence becomes the most challenging issue of state-of-the-art FAS research.

\begin{figure}[t]
    \centering
    \vspace{-0.8em}
    \includegraphics[width=0.47\textwidth]{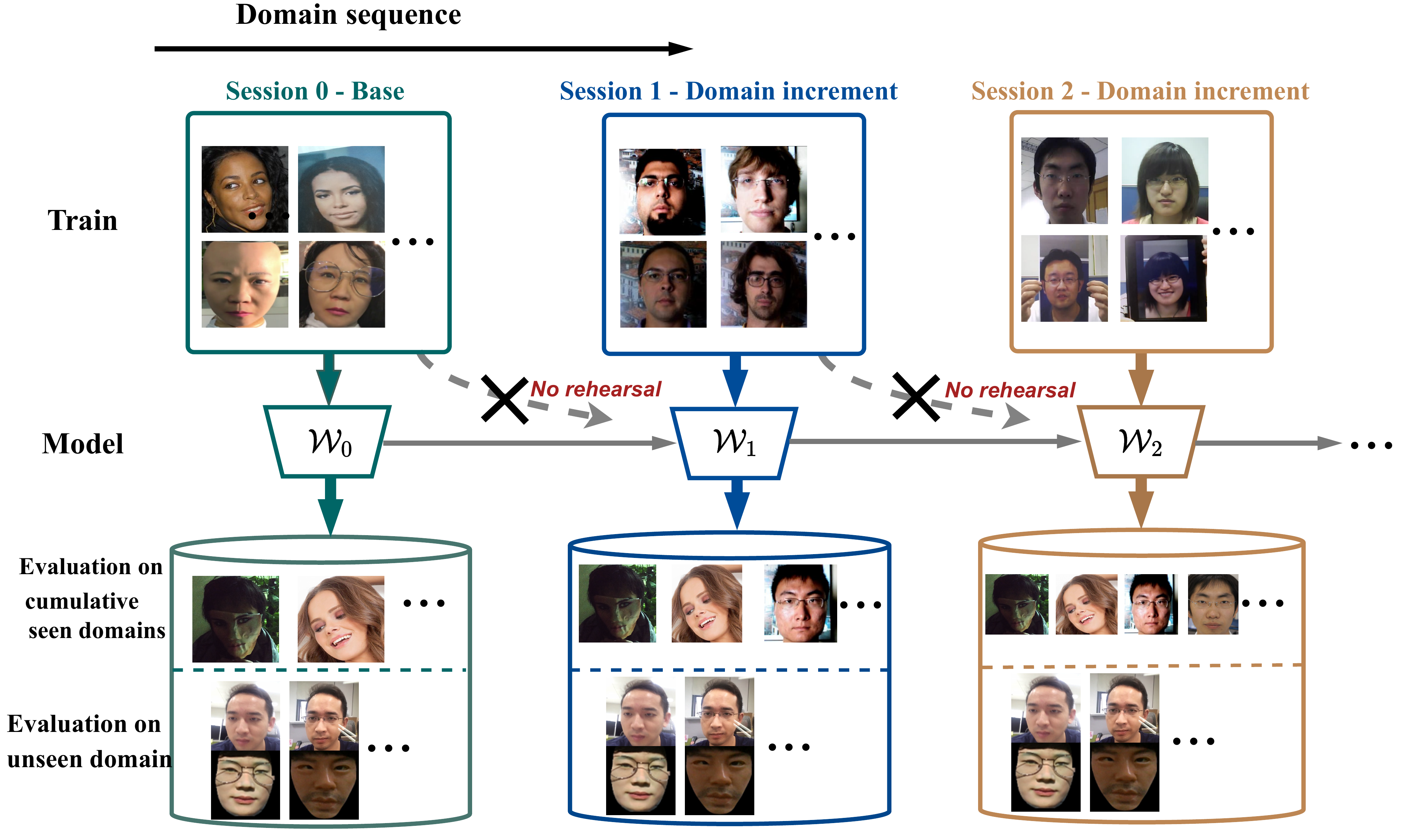}
    \vspace{-0.4em}
    \caption{The rehearsal-free DCL consists of a sequence of learning sessions. The FAS model is initially trained on a large-scale base domain and then continually adapts to new domains in the following continual sessions. For each continual session, only a few data of new domain is available for training and previous data is NOT available. After the DCL, the model is tested on all previous domains and extra unseen domains.}
    \label{fig:my_label}
    \vspace{-1em}
\end{figure}

To tackle the cross-domain problem, domain generalization~\cite{MADDG-CVPR-2019,SSDG-CVPR-2020} and adaption~\cite{li2018unsupervised,huang2022adaptive} techniques for FAS have been extensively studied in recent years. Domain generalization-based methods aim to develop a generalized FAS model with training data from multiple source domains. Despite improving the generalization to some extent, they are still far from satisfaction in unseen domains. Besides, domain adaptation-based methods utilize target domain data for model adaptation. Although the target domain performance can be significantly improved, the benefit is at the cost of expensive target data collection. Moreover, it is even impractical to collect sufficient data at a static point of time since the domain shifts are caused by constantly changing factors like illuminations and attack types.

In real-world scenarios, the deployed FAS systems constantly encounter new data from various domains. The new data will be collected and become available for model training gradually. Completely retraining a model from scratch with old and new data have both efficiency and privacy issues. Although fine-tuning the base model with the new data only is more efficient, the past knowledge will be overwritten after fine-tuning, and the performance on previous data decreases dramatically, \textit{i.e.} catastrophic forgetting \cite{perez2020learning}. To adapt models efficiently, continual learning methods for FAS have been proposed in recent works \cite{perez2020learning, iccv2021_dcl_fas}. To alleviate the catastrophic forgetting, both methods \cite{perez2020learning, iccv2021_dcl_fas} utilize replay buffers to store previous data for rehearsal while fine-tuning with new data. However, the use of replay buffers causes extra storage burdens. Even worse, the previous data is not always available for storage and transfer since face data contains identity information.

In this work, we tackle the FAS problem under the rehearsal-free Domain Continual Learning (DCL) setting. Unlike existing work \cite{iccv2021_dcl_fas}, where the FAS model is expected to learn sequentially from data of novel attack types, the aim of our work is enabling the FAS model continually evolve with the data from constantly varying domains. Due to efficiency and privacy issues, previous data is not allowed to be stored and accessed for rehearsal and only a few (low-shot) new data available for continual learning, which are different from previous works \cite{perez2020learning, iccv2021_dcl_fas}. \rz{We first evaluate the baseline method under the DCL setting and obtain the below interesting observations from experiments: catastrophic forgetting usually occurs when the new coming data dataset has large domain gaps from previous ones, and a model with better unseen domain generalization performance usually forgets less previous domain knowledge. Motivated by above the observations, we propose to address the DCL-FAS problem from the aspect of generalization.}
\par
\rz{During continual sessions, a small amount of data could lead to overfitting, bring poor generalization performance and catastrophic forgetting. To update models continually and efficiently, we introduce Efficient Parameter Transfer Learning (EPTL) paragdim for the DCL-FAS and utilize Adapters \cite{houlsby2019parameter,huang2022adaptive} for Vision Transformer (ViT) \cite{dosovitskiy2020image}. By using the adapters, \cite{houlsby2019parameter}, ViT models can be efficiently adapted even with low-shot training data. However, we find that vanilla adapters consisting of linear layers cannot satisfy the need of extracting fine-grained features for the FAS task. Hence, we replace the vanilla Linear Adapter with our proposed Dynamic Central Difference Convolutional Adapter (DCDCA), which empowers ViT with image-specific inductive bias by convolution and extracts fine-grain features with adaptive central difference information \cite{CDCN-CVPR-2020}. Unlike \cite{CDCN-CVPR-2020} where the ratio of central difference information is fixed for all layers, the ratio in our designed DCDCA is self-adaptive to new data domains, which is more suitable in the DCL setting. Besides, to further improve the generalization performance, we optimize DCDCA with contrastive regularization, and reduce forgetting during optimization by our proposed proxy prototypes the contrastive regularization (PPCR). Without the access previous data, our PPCR utilize previous data knowledge extracted from the class centroids of previous tasks, which are approximated by model weights of the fully-connected layers, instead of previous data.}

Our contributions include: \textbf{\textit{1)}} We formulate and tackle the FAS problem in a more practical scenario: low-shot and rehearsal-free Domain Continual Learning (DCL). In each continual learning session, only a few new data is available for training and no previous data is accessible; \textbf{\textit{2)}} We design the Dynamic Central Difference Convolutional Adapter (DCDCA) to efficiently adapt ViT-based models in continual domains and capture intrinsic live/spoof cues; \textbf{\textit{3)}} We propose the Proxy Prototype Contrastive Regularization (PPCR) to further improve the generalization and alleviate the forgetting of FAS models during rehearsal-free DCL; \textbf{\textit{4)}} We design two practical protocols to evaluate both anti-forgetting and generalization capacities of FAS models under DCL settings, with up to 15 public datasets covering both 2D and 3D attacks. We find that the proposed DCDCA and PPCR can significantly improve generalization while forgetting less over baselines on these two DCL protocols.  

\begin{figure*}
\vspace{-1mm}
\centering
\includegraphics[width=0.88\textwidth]{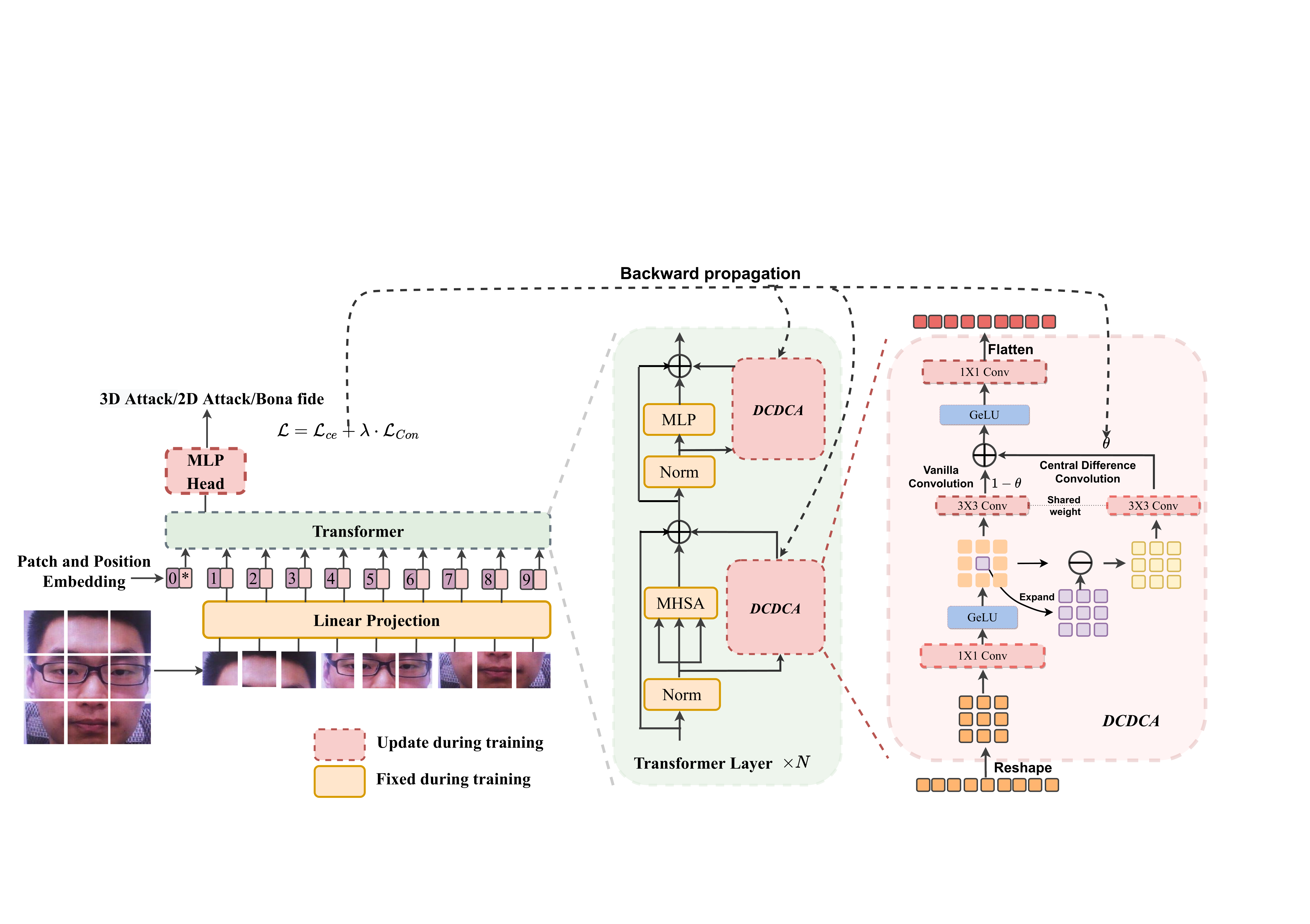}
\vspace{-2mm}
\caption{The architecture of the proposed ViT-DCDCA. The Dynamic Central Difference Convolutional Adapter (DCDCA) is able to extract the fine-grained central difference information for intrinsic live/spoof representation. Only the DCDCA and the MLP head are updated during training. `MHSA' and `MLP' denote multi-head self-attention and multi-layer perceptron, respectively.}\label{fig:DCDCA}
\vspace{-0.9em}
\end{figure*}

\vspace{-1em}
\section{Related works}
\subsection{Cross-Domain Face Anti-Spoofing}
Due to the powerful representation ability of deep neural networks, deep learning based FAS methods gradually surpass and replace the traditional methods based on handcrafted features \cite{ramachandra2017presentation, yu2022deep, kong2022digital}. Recently, plenty of domain generalization and adaptation techniques have been proposed to improve the cross-domain FAS performance. 

Domain generalization-based methods~\cite{li2018learning,wang2020cross, DRDG_IJCAI21} aim to improve the generalization ability of FAS models by learning generalized feature representations with training data of multiple data domains. Li \textit{et al.} \cite{li2018learning} proposed to learn feature representations via domain alignment, which minimizes the distance between feature distributions of different data domains. To improve the generalization ability, disentangled representation learning techniques have been used to extract domain-dependent features \cite{wang2020cross, VSA_MM21}. To deal with the biased distribution of different data domains, Liu \textit{et al.} \cite{DRDG_IJCAI21} proposed to re-weight the relative importance between different samples. Besides, meta-learning concepts have been extensively used in FAS to improve the generalization performance via elaborate meta-tasks \cite{LMM_AAAI20, RFM_AAAI20, ANRL_MM21, MetaTeacher-TPAMI-2021, metapattern}. Considering that manually partitioning training data into different domains is expensive and sub-optimal, Chen \textit{et al.} \cite{chen2021generalizable} proposed a method dividing training data from a mixture of data domain automatically. Although these methods could learn more generalized FAS models without any target domain data, their performance is not always satisfactory, especially when there are large domain gaps between source and target domains. 

Domain adaptation \cite{li2018unsupervised, wang2020unsupervised, wang2020cross,LMM_AAAI20, huang2022adaptive,qin2020one,li2022one} based methods help further improve the performance of FAS models with target domain data for adaptation. To avoid the expense of target domain data annotation, unsupervised domain adaptation methods have been proposed to improve the cross-domain performance of FAS with some unlabelled target domain data \cite{li2018unsupervised, wang2020unsupervised, wang2020cross}. Considering the expense of target data collection, few-shot learning concept has been incorporated into the domain adaptation methods \cite{LMM_AAAI20, huang2022adaptive}. Since the data collection of genuine face samples is easier and cheaper than spoofing ones, recent works \cite{qin2020one, li2022one} proposed methods that only use genuine face samples of target domain for one-class domain adaptation. However, most existing methods either require using source domain data during the adaptation or suffer from catastrophic forgetting of the source domain after adaptation.

\vspace{-0.5em}
\subsection{Continual Learning}
Due to uncontrollable factors such as the changes in the environment and the emergence of novel attack types, the FAS systems will always encounter testing examples that are from unseen data distributions. The FAS systems are thereby expected to have the ability to continually learn from newly collected data and adapt themselves like humans. Naively fine-tuning neural network models with new data usually overwrites the knowledge learned from previous data \cite{kemker2018measuring, maltoni2019continuous}. Continual learning \cite{parisi2019continual} aims to alleviate catastrophic forgetting with elaborate regularization \cite{lwf, ewc} or memory replay \cite{shin2017continual, kemker2017fearnet}. Recently, some continual learning frameworks for FAS have been proposed \cite{perez2020learning, iccv2021_dcl_fas}, which alleviate the forgetting of FAS models with the help of replay buffers. However, the use of a replay buffer cause storage inefficiency and privacy issues since facial image as biometric information is sensitive for storage and transfer. Different from existing works, we propose the first rehearsal-free DCL framework for FAS without using replay buffers to store previous data.

\vspace{-0.2cm}
\section{Methodology} \label{sec-method}
\subsection{Dynamic Central Difference Convolutional Adapter}
Given the observation from experiments that a model that generalizes well can usually have less catastrophic forgetting (see Sec.~\ref{sec:Analysis1} and ~\ref{sec:Analysis2}), to achieve the goals of DCL-FAS: generalize more and forget less, we propose Dynamic Central Difference Convolutional Adapter (DCDCA), which adapts ViT with dynamic central difference information during the continual learning.

\vspace{0.3em}
\noindent\textbf{Fine-tuning ViT with adapter.}\quad
ViT \cite{dosovitskiy2020image} consists of a stack of transformer blocks, and each block comprises a Multi-Head Self Attention (MHSA) layer and Multilayer Perceptron (MLP) layers to extract features. By ignoring the skip connection and Norm layer, the inference procedure can be expressed as
\vspace{-0.6em}
\begin{equation}
\vspace{-0.4em}
     out = \text{MLP}_\mathcal{W}(\text{MHSA}_\mathcal{W}(x)),
\end{equation}
where $x$ is the input token, and $\mathcal{W}$ represents the parameters of the transformer, $out$ is the output token. Although ViT has strong feature representation capability, there are a large number of parameters to update when fine-tuning the ViT to a downstream task. It usually requires a large amount of data and training time. Recent studies of parameter-efficient transfer learning (PETL) on transformers \cite{huang2022adaptive,vpt,convpass} show that inserting adapter layers is an efficient way to fine-tune ViT. Such PETL paradigm is named as ViT-Adapter.
Vanilla ViT-Adapter usually has extra linear layers $\mathcal{A}$ inserted into transformer layers. As such, the inference of a ViT-Adapter is expressed as 
\vspace{-0.4em}
\begin{equation}
\vspace{-0.4em}
     out = \mathcal{A}(\text{MLP}_\mathcal{W}( \mathcal{A} (\text{MHSA}_\mathcal{W}(x))).
\end{equation}
When using a ViT model for a new downstream task, parameters of the pretrained ViT backbone ($\mathcal{W}$) are fixed, and only the parameters of the inserted adapter layers ($\mathcal{A}$) are updated. As $\mathcal{A}$ takes up a small ratio of parameters compared to the entire ViT, PETL requires only a small amount of training data and applies to DCL-FAS where a limited new domain data is available in the continual sessions.

\vspace{0.3em}
\noindent\textbf{Fine-tuning with DCDCA.}\quad
Inspired by central difference convolution (CDC) \cite{CDCN-CVPR-2020,NASFAS-TPAMI-2020} that extracts more robust feature representation for FAS by integrating local descriptors with convolution operation, we propose the Dynamic Central Difference Convolution Adapter (DCDCA) to introduce the locality inductive bias for ViT and extract fine-grained information with CDC \cite{CDCN-CVPR-2020}. As illustrated in Fig. \ref{fig:DCDCA}, the DCDCA is embedded in the ViT backbone as a residual bottleneck connection \cite{convpass,ResNet}. During the continual learning, only the DCDCA and the classification MLP head are updated, while the other pretrained layers are fixed.

Specifically, 2D convolutional layers inside the DCDCA are utilized to provide the locality inductive bias. To fit the convolution operation, the 1D flattened image token from the ViT backbone is reshaped back to a 2D structure for processing. Then, the reshaped 2D token is forwarded to a stack of convolutional layers for feature extraction. To extract features for subtle live/spoof discrimination, we use CDC to extract fine-grained contextual info from neighbor visual tokens. The output $y(p_0)$ is defined as 
\vspace{-0.5em}
\begin{equation}\label{eq-cdc}\small
\vspace{-0.4em}
\begin{aligned}
    y(p_0) =& \theta \underbrace{\sum_{p_n\in \mathcal{R}}\omega(p_n)\cdot (x(p_0+p_n) - x(p_0))}_{\text{central difference convolution}} + \\ 
     &  \underbrace{(1 - \theta) \sum_{p_n\in \mathcal{R}} \omega(p_n)\cdot x(p_0+p_n)}_{\text{vanilla convolution}}, 
\end{aligned}
\end{equation}
where $\omega$ is the convolutional kernel, $p_0$ is the center token of a 2D token map and $\mathcal{R}$ denotes the neighbor tokens around the token $p_0$. $\theta$ is the ratio of central difference information, which is empirically set as $0.7$ for all layers in \cite{CDCN-CVPR-2020}. However, using a united and fixed $\theta$ in all CDC layers is sub-optimal for DCL-FAS from two perspectives. First, the proportion of central difference information in features should be layer-specific because the semantic information and grain fineness of features are different among hierarchical layers. Second, in the continual learning scenario, the data domains change dynamically thus the contribution of central difference cues should be dynamically adapted as well. Therefore, we parameterize the $\theta$ of DCDCA as learnable variables that are self-adaptable to different layers and continual learning sessions.

To increase generalization capability, we propose to treat Eq.~\ref{eq-cdc} as a type of feature transformation \cite{FWT}, which transforms vanilla convolution feature with a scaling factor $\Theta$ sampled from a learnable Gaussian Distribution $\text{N}(\mu, \sigma^2)$. Since the sampling would stop the gradient backward propagation, we utilize the re-parameterization skill of Gaussian distribution that
\vspace{-0.4em}
\begin{equation}
\vspace{-0.4em}
    \begin{aligned}
     \Theta\sim \text{N}(\mu, \sigma^2) \iff \Theta=\mu+\sigma\cdot\epsilon, \epsilon\sim \text{N}(0, 1)
    \end{aligned}.
\end{equation}
During training, $\mu$ and $\sigma$ are updated to sample $\Theta$, and we use $\theta=\text{Sigmoid}(\Theta)$ in Eq~\ref{eq-cdc}, where $\text{Sigmoid}$ is used to constrain the output in $[0,1]$. During testing, randomness is removed, and $\Theta=\mu$. We also compare our domain-aware dynamic $\theta$ estimation method with other learnable $\theta$ strategies in \textit{Appendix}.

\begin{figure}
\centering
\includegraphics[width=0.9\linewidth]{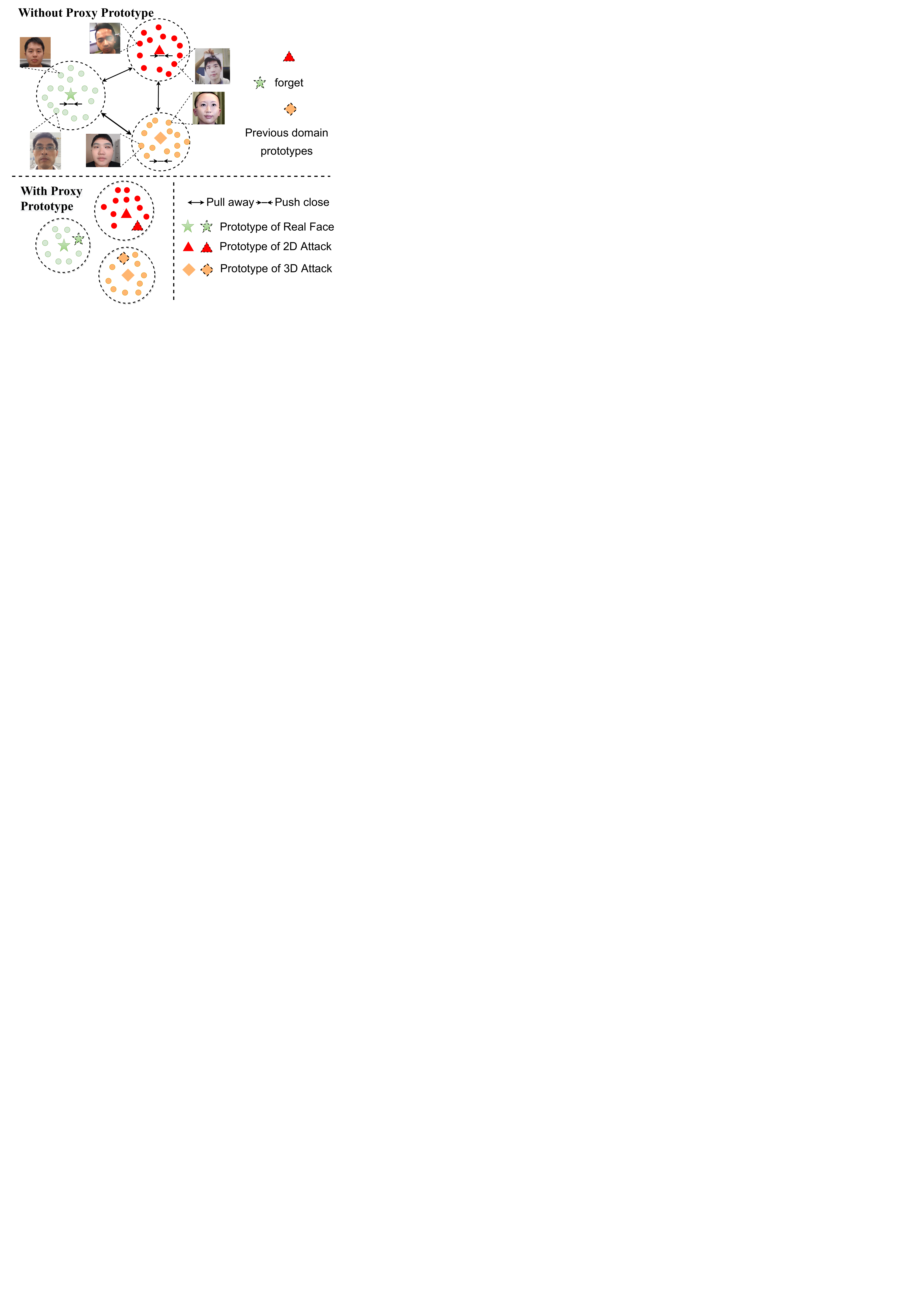}
\vspace{-0.6em}
\caption{Illustration of Proxy Prototype Contrastive Regularization (PPCR). The top shows when learning on a new domain without prototypes, the new features might shift away from the previous prototypes, and previous knowledge is forgotten. The bottom shows our PPCR regularizes the new features to be clustered near the previous prototypes and forget less previous knowledge.} \label{fig:ppcr}
\vspace{-0.8em}
\end{figure}

\subsection{Proxy Prototype Contrastive Regularization}
\vspace{-0.2em}
To learn more generalized models, supervised contrastive loss \cite{SupCon} is adapted for network optimization. Considering that the distributions of real face samples are relatively similar than spoofing ones \cite{SSDG-CVPR-2020}, all real face samples are regarded as one cluster while spoofing face samples are divided into 2D attacks and 3D mask attacks. Therefore, the loss for optimization is expressed as
\vspace{-0.2em}
	\begin{equation}\label{eq-supcon}\small
	\vspace{-0.2em}
	    \begin{aligned}
	        \mathcal{L}_{Con} &=\mathcal{L}_{Con}^{\mathcal{C}^1}+ \mathcal{L}_{Con}^{\mathcal{C}^{2}} + \mathcal{L}_{Con}^{\mathcal{C}^{3}}, \\
	        \mathcal{L}_{Con}^{\mathcal{C}^1} &= \sum_{i\in\mathcal{C}^1} \frac{-1}{|\mathcal{C}^1|}\sum_{j\in \mathcal{C}^1, j\ne i}log \frac{\text{exp}(z_i\cdot z_j)}{\sum_{a\in\mathcal{C}^{2}\cap\mathcal{C}^{3}} \text{exp}(z_i, z_a)}, \\
	        \mathcal{L}_{Con}^{\mathcal{C}^{2}} &= \sum_{i\in\mathcal{C}^{2}} \frac{-1}{|\mathcal{C}^{2}|}\sum_{j\in \mathcal{C}^{2}, j\ne i}log \frac{\text{exp}(z_i \cdot z_j)}{\sum_{a\in\mathcal{C}^1\cap\mathcal{C}^{3}}  \text{exp}(z_i, z_a)}, \\
	        \mathcal{L}_{Con}^{\mathcal{C}^{3}} &= \sum_{i\in\mathcal{C}^{3}} \frac{-1}{|\mathcal{C}^{3}|}\sum_{j\in \mathcal{C}^{3}, j\ne i}log \frac{\text{exp}(z_i \cdot z_j)}{\sum_{a\in\mathcal{C}^1\cap\mathcal{C}^{2}}  \text{exp}(z_i, z_a)} 
	    \end{aligned}
\end{equation}
where $\mathcal{C}^1$, $\mathcal{C}^{2}$, and $\mathcal{C}^{3}$ denote the set of sample indices of real face, 2D attack and 3D attack examples respectively, $|\mathcal{C}^k|$ denotes the number of samples in $\mathcal{C}^k$, $z_i$ denotes the feature from the last transformer layer of a sample $i$. 

\par 
After the features of the same class are aligned and clustered, the clusters' centroids are set as prototypes. When continually learning with a new data domain, FAS model will forget the previous knowledge if the features of new domain data are far away from the old prototypes as illustrated in the upper part of Fig.~\ref{fig:ppcr}. Recent research of source-free model transfer \cite{liang2020we} shows that model weight can provide knowledge of the source training data and the linear classifier $\mathbb{f}$ of a model is equivalent to the prototype in supervised contrastive learning \cite{eccv2022-sourcefree-liu}. Therefore, we propose the Proxy Prototype Contrastive Regularization (PPCR) to reduce forgetting during continual learning without accessing previous data. We set proxy prototypes $\mathbb{f}$ as the anchors in contrastive training and regularize clustering with previous prototypes to make the previous knowledge less forgotten, as illustrated in the bottom part of Fig.~\ref{fig:ppcr}. We define the linear classifier weight as $\mathbb{f}=\{f^1, f^2, f^3\}$, where $f^1$, $f^2$, and $f^3$ are the weights and the proxy prototypes of the classes of real face, 2D attack, and 3D attack, respectively. Then, we define the final loss for optimization as
\begin{equation}\label{eq-overall}
    \mathcal{L} = \mathcal{L}_{CE} + \lambda \mathcal{L}_{Con},
\end{equation}
where $\mathcal{L}_{CE}$ is the cross-entropy loss, and $\lambda$ is a constant scaling factor to balance two terms.  Finally, the overall algorithm for the DCL-FAS with our proposed PPCR is described in Algorithm~\ref{algo}.

\begin{table*}[tbp]
  \centering
     \caption{Illustration of Proctocol-1 and Protocol-2. $\dag$ denotes the dataset contains 2D attack and $\ddag$ means the dataset contains 3D attack.} 
     \label{tab:protocol}%
       \resizebox{0.95\textwidth}{!}{
    \begin{tabular}{|c|cccccccccc|}
    \hline
    Base & \multicolumn{10}{c|}{Celeba-Spoof$^{\dag}$, SiW$^{\dag}$, HiFI Mask$^{\ddag}$ } \bigstrut\\
    \hline
    Session \newline{}ID & \multicolumn{1}{c|}{1} & \multicolumn{1}{c|}{2} & \multicolumn{1}{c|}{3} & \multicolumn{1}{c|}{4} & \multicolumn{1}{c|}{5} & \multicolumn{1}{c|}{6} & \multicolumn{1}{c|}{7} & \multicolumn{1}{c|}{8} & \multicolumn{1}{c|}{9} & 10 \bigstrut\\
    \hline
    Protocol\newline{} 1 & \multicolumn{1}{c|}{REPLAY-ATTACK$^{\dag}$} & \multicolumn{1}{c|}{CASIA\newline{}-FASD$^{\dag}$} & \multicolumn{1}{c|}{MSU\newline{} MFSD$^{\dag}$} & \multicolumn{1}{c|}{HKBU \newline{}MarV2$^{\ddag}$} & \multicolumn{1}{c|}{OULU-NPU$^{\dag}$} & \multicolumn{1}{c|}{CSMAD$^{\ddag}$} & \multicolumn{1}{c|}{CASIA-SURF$^{\dag}$} & \multicolumn{1}{c|}{WFFD$^{\ddag}$} & \multicolumn{1}{c|}{WMCA$^{\dag \ddag}$} & \multicolumn{1}{c|}{CASIA-3DMASK $^{\ddag}$} \bigstrut\\
    \hline
    Protocol \newline{}2 & \multicolumn{1}{c|}{CASIA-3DMASk$^{\ddag}$} & \multicolumn{1}{c|}{WMCA$^{\dag\ddag}$} & \multicolumn{1}{c|}{WFFD$^{\ddag}$} & \multicolumn{1}{c|}{CASIA\newline{}-SURF$^{\dag}$} & \multicolumn{1}{c|}{CSMAD$^{\ddag}$} & \multicolumn{1}{c|}{OULU\newline{}-NPU$^{\dag}$} & \multicolumn{1}{c|}{HKBU \newline{}MarV2$^{\ddag}$} & \multicolumn{1}{c|}{MSU\newline{} MFSD$^{\dag}$} & \multicolumn{1}{c|}{CASIA\newline{}-FASD$^{\dag}$} & \multicolumn{1}{c|}{REPLAY-ATTACK$^{\dag}$} \bigstrut\\
    \hline
    Unseen & \multicolumn{10}{c|}{ROSE-YOUTU$^{\dag}$, CeFA$^{\dag \ddag}$} \bigstrut\\
    \hline
    \end{tabular}%
    }
  \vspace{-1.0em}
\end{table*}%

\begin{algorithm}[t]
\caption{Domain Continual Face Anti-Spoofing with PPCR} \label{algo}
        \SetAlgoLined
        An ImageNet pretrained ViT backbone  $\mathcal{W}=\{\mathcal{W}_b, \mathbb{f}\}$; \\
        Insert DCDCA modules $\mathcal{A}$ to the backbone; \\
        Train the network on the base dataset, and only $\mathcal{A}$and $\mathbb{f}$ are updated; \\
        \For{$t=1$ to $T$}{
            Clone and detach $f^1$, $f^2$, and $f^3$ from $\mathbb{f}$;
            
            \While{Session $t$ not finished}{
                Sample a batch of data $X^b$
                Conduct inference on $X^b$ and sort out the features of samples into $\mathcal{C}^1$, $\mathcal{C}^2$, and $\mathcal{C}^3$;
                
                Include proxy prototypes: $\mathcal{C}^1 = \mathcal{C}^1\cup f^1$,  $\mathcal{C}^2 = \mathcal{C}^2\cup f^2$, $\mathcal{C}^3 = \mathcal{C}^3\cup f^3$;
                
                Calculate loss based on Eq.~\ref{eq-overall};
                
                Conduct backward propagation to update $\mathcal{A}$ and $\mathbb{f}$}   }
                
                Output: the optimized $\mathcal{A}$ and $\mathbb{f}$
                
\end{algorithm}

\section{Experiment}
\subsection{Protocols for DCL-FAS}
To establish DCL-FAS setting, we construct two practical continual learning protocols based on the RGB data of 15 publicly available datasets: IDIAP REPLAY-ATTACK \cite{idiapREPLAY-ATTACK}, CASIA-FASD \cite{DB-CASIAFASD}, MSU MFSD \cite{wen2015face}, HKBU MARsV2\cite{HKBU_Marv2}, OULU-NPU\cite{OULU_NPU_2017}, CSMAD\cite{CSMAD}, CASIA-SURF\cite{CASIA-SURF}, WFFD\cite{WFFD}, WMCA\cite{george2019biometric}, CASIA-SURF 3DMASK (CASIA-3DMASK) \cite{NASFAS-TPAMI-2020}, ROSE-YOUTU \cite{li2018unsupervised}, CASIA-SURF CeFA (CeFA) \cite{CeFA}, CelebA-Spoof \cite{CelebA-Spoof}, CASIA-SURF HiFiMask \cite{HIFIMASK} and SiW \cite{FAS-Auxiliary-CVPR-2018}. The details about the above datasets can be found in the \textit{Appendix}.

\begin{figure*}
    \centering
    \includegraphics[width=\textwidth]{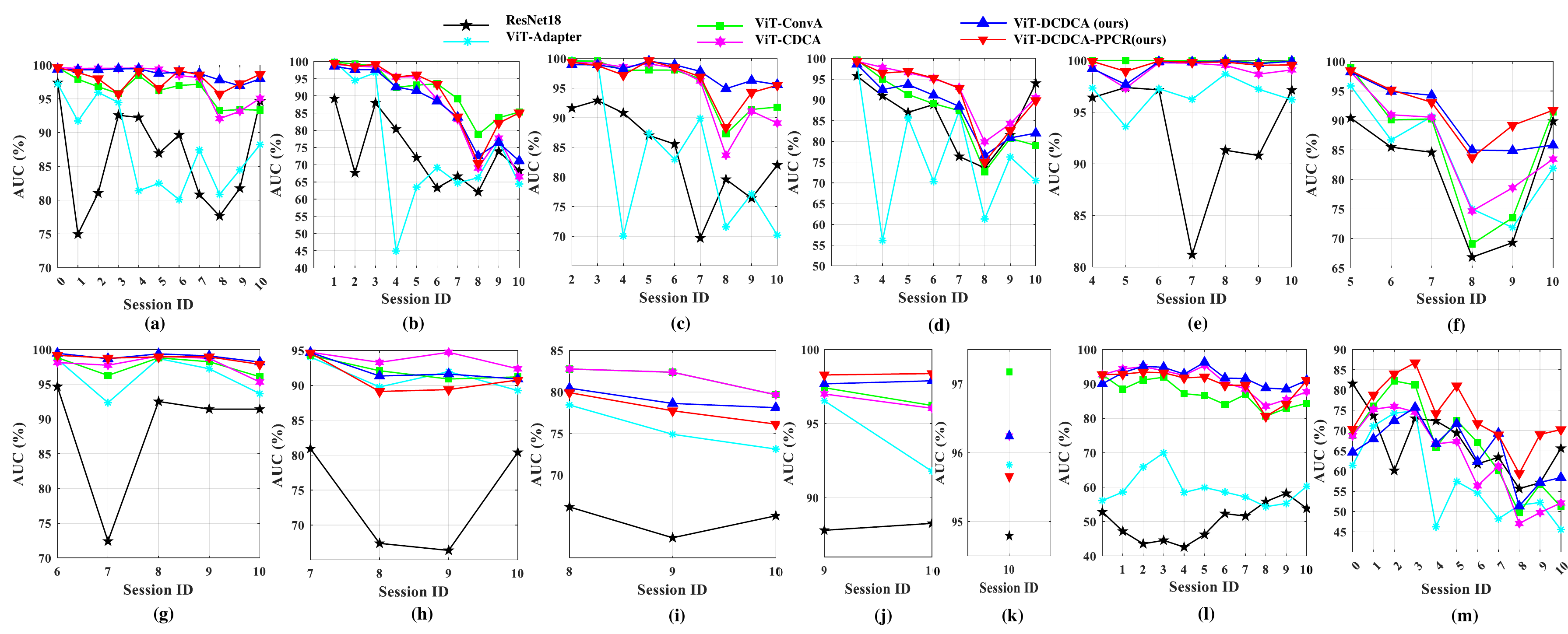}
    \vspace{-1.8em}
    \caption{Performance of proposed Protocol-1 with the used architectures. (a)-(k) show the models performance on $\mathcal{D}_0$ to $\mathcal{D}_{10}$ in different sessions. (l) and (m) show the testing performance of unseen data domains ROSE-YOUTU and CeFA in different sessions.}
    \label{fig:P1}
     \vspace{-0.5em}
\end{figure*}

\noindent\textbf{Shared configuration of the starting session.} We first describe the shared configuration for Protocol-1 and Protocol-2. In practical development, to train a base model (base session, $t=0$), a large-scale of data including various 2D and 3D attack samples is often collected as the base dataset. As such, we combine SiW, Celeba-Spoof, and HiFiMask datasets as the base dataset as they contain large amount of data. Protocol-1 and Protocol-2 share the same datasets for base model training.
When adapting the base model for a new domain, we consider the prompt development where the model should be adapted efficiently to a small collection of data. Hence, we set up a low-shot condition in continual session $t>0$ by randomly extracting 50 frames of real face examples and 50 frames of spoofing attack examples from the training partition of the new dataset. At the testing stage, all samples of the testing partition of the dataset are used for model evaluation.

\begin{figure*}
    \centering
    \vspace{-0.8em}
    \includegraphics[width=\textwidth]{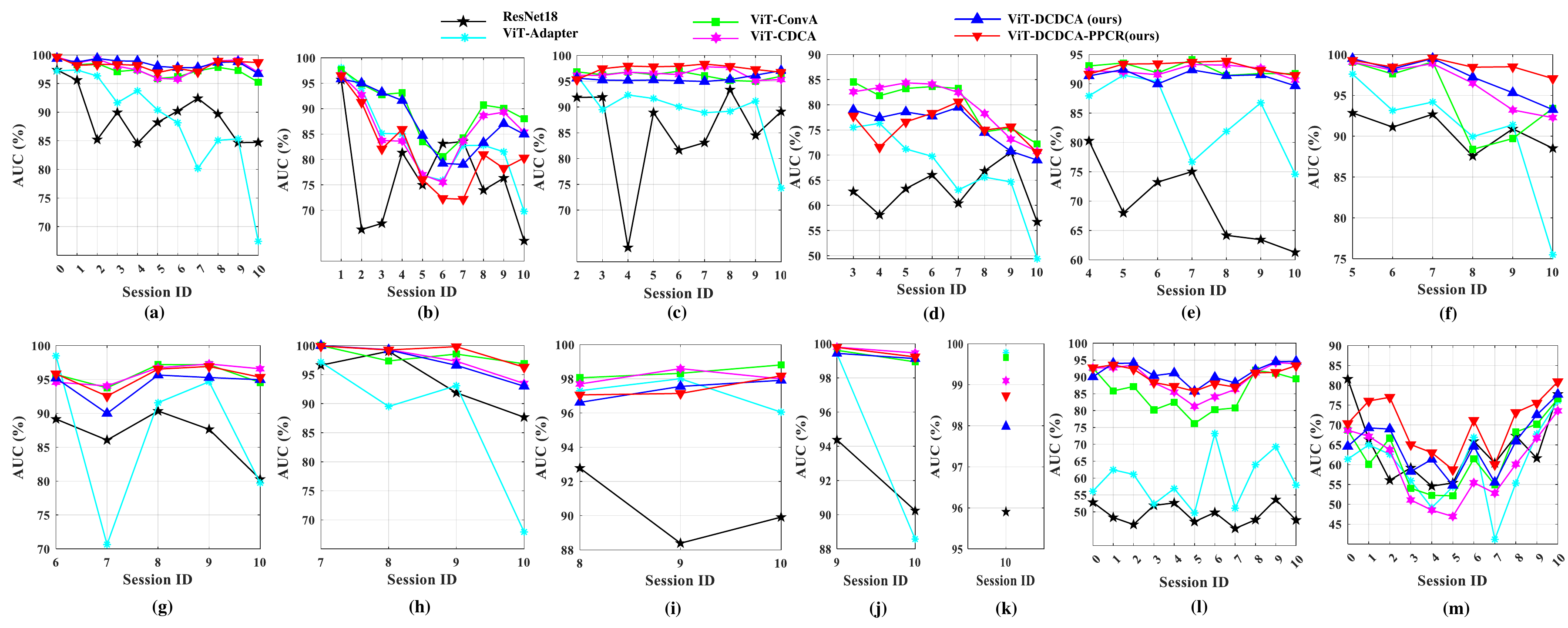}
    \vspace{-1.5em}
    \caption{Performance of proposed Protocol-2 with the used architectures. (a)-(k) show the models performance on $\mathcal{D}_0$ to $\mathcal{D}_{10}$ in different sessions. (l) and (m) show the testing performance of unseen data domains ROSE-YOUTU and CeFA in different sessions. }
    \label{fig:P2}
    \vspace{-0.5em}
\end{figure*}

\textbf{Protocol-1} simulates the scenario where a base model is adapted as new data domains emerge. As such, from Session 1 to 10, we arrange the incoming domain sequence according to the release years of the public datasets by the ascending orders (from old to new), as shown in Table~\ref{tab:protocol}. 
\textbf{Protocol-2} simulates the scenario where a base model needs to be compatible to data from old devices. As such, Protocol-2 is set up by reverting Protocol-1 (from old to new), as shown in Table~\ref{tab:protocol}. For both protocols, we use ROSE-YOUTU and CeFA datasets as unseen datasets as ROSE-YOUTU has diverse 2D attack samples and CeFA includes diverse 2D and 3D attack samples. The model's unseen domain generalization performance is evaluated on these two datasets after the training of each session.

\vspace{0.3em}
\noindent\textbf{Evaluation metrics.}
We first introduce the notations we used before describing the evaluation metrics. The session ID is denoted by $t$, and $t=0$ is the base session, and there are total $T+1$ sessions with $T$ incremental sessions. In Session $t$, the training dataset is denoted as $\mathcal{D}_t$. After the training in Session $t$, the model status is denoted as $\mathcal{W}_t$. The model $\mathcal{W}_t$ is evaluated on the testing data of previously seen datasets $\mathcal{D}_e (e\leq t)$, and the corresponding performance is denoted as $R_{t,e}$.

Motivated by \cite{lopez2017gradient}, we use Area Under the Receiver operating characteristic Curve (AUC) to define mean Average AUC ($mAA$), mean Accumulative Backward Transfer of AUC ($mABT$), and Average unseen domain Generalization AUC ($mAGA$) as

\begin{equation}
\vspace{-0.6em}
\begin{aligned}
     mAA &\triangleq\frac{1}{T+1}\sum_{t=0}^T (\frac{1}{T-t+1} \sum_{i=t}^T R_{t,i}),\\
     mABT &\triangleq\frac{1}{T}\sum_{t=0}^{T-1} (\frac{1}{T-t}\sum_{i=t+1}^{T}R_{i,t}- R_{t,t}),\\
    AGA_u &\triangleq \frac{1}{T+1} \sum_{t=0}^T R^u_{t},\\ mAGA&\triangleq\frac{1}{2}(AGA_{1}+AGA_{2}),
    \end{aligned}
\vspace{0.1em}
\end{equation}

where $R_t^u$ means testing $\mathcal{W}_t$ on a unseen domain $u$, $AGA_1$ and $AGA_2$ denotes the performance on ROSE-YOUTU and CeFA respectively.
The $mAA$ (the higher, the better) measures the average intra-domain AUC performance, and $mAGA$ (the higher, the better) evaluates the average cross-domain performance in the unseen domains.
Our designed $mABT$ is similar to Backward Transfer (BWT) \cite{lopez2017gradient}, and high negative values of $mABT$ mean more forgetting. While BWT merely considers the final session to measure forgetting, our $mABT$ considers all intermediate sessions due to practical concerns. In practical development, it is undetermined when a new domain session $t+1$ comes and the model should be deployed after a session $t$. Therefore, our $mABT$ involves the results of intermediate sessions to measure the average forgetting.

\vspace{-1mm}
\noindent\textbf{Implementation details.}
We conduct experiments with ResNet18. Besides, we use `vit\_base' \cite{dosovitskiy2020image} as the backbone for different adapters in our work. ViT-Adapter denotes inserting vanilla Linear layers after the MHSA and MLP blocks of the ViT backbone. We insert our proposed DCDCA to ViT and derive ViT-DCDCA, as shown in Fig.~\ref{fig:DCDCA}. If we fix $\theta=0.7$, ViT-DCDCA degrades to ViT-CDCA (Central Difference Convolutional Adapter). If we fix $\theta=0.0$, there is no central difference and ViT-CDCA becomes ViT-ConvA (Convolutional Adapter). In the training of the base model ($t=0$), the number of the epoch is 10, the batch size is 64. In each continual session ($t>0$), the model is trained for 100 epochs and the batch size is 20. $\lambda=0.05$ is used for the PPCR. For all training, the Adam optimizer is used with an initial learning rate 0.0001.

\begin{table}[tbp]
  \centering
  \caption{Compare different architectures on Protocol-1 and Protocol-2. Better performance (\%) in comparison is in \textbf{bold}.} 
  \vspace{-0.1em}\label{tab:comparebackbone}%
      \resizebox{1.0\linewidth}{!}{
    \begin{tabular}{l|ccccccc}
    \hline
    \multicolumn{1}{c|}{\multirow{2}[4]{*}{Architecture}} & \multicolumn{3}{c}{Protoco-1} &       & \multicolumn{3}{c}{Prococol-2} \bigstrut\\
\cline{2-4}\cline{6-8}          & $mAA$ & $mABT$& $mAGA$  &       & $mAA$ & $mABT$& $mAGA$ \bigstrut\\
    \hline
    ResNet18 & 83.02 & -7.19 & 58.29 &       & 83.28 & -7.36 & 56.76 \bigstrut[t]\\
    ViT-Adapter & 86.72 & -9.87 & 58.72 &       & 87.93 & -7.74 & 59.53 \\
    ViT-ConvA & \textbf{93.29} & -4.24 & 76.73 &       & \textbf{94.65} & -2.28 & 73.79 \\
    ViT-CDCA (ours) & 93.06 & -4.08 & 77.01 &       & 94.42 & -1.89 & 74.48 \\
    ViT-DCDCA(ours) & 93.23 & \textbf{-3.59} & \textbf{78.70} &       & 93.79 & \textbf{-1.82} & \textbf{78.09} \bigstrut[b]\\
    \hline
    \end{tabular}%
    }
    \vspace{-0.5cm}
\end{table}%

\vspace{-0.2em}
\subsection{Qualitative Analysis}
\vspace{-0.1em}
\label{sec:Analysis1}
Fig.~\ref{fig:P1} and Fig.~\ref{fig:P2} show the results of Protocol-1 and Protocol-2, respectively, and all the detailed numbers can be found in the \textit{Appendix}. In Fig.~\ref{fig:P1} and Fig.~\ref{fig:P2}, ViT-DCDCA-PPCR means the model is trained by our PPCR algorithm, while the others are trained by using finetuning with $\mathcal{L}_{CE}$.  For both figures, the sub-figures (a)-(k) show testing models on the data domains from $\mathcal{D}_0$ to $\mathcal{D}_{10}$. For example, Fig.~\ref{fig:P1}(a) show the results of testing models $\mathcal{W}_i$ on the base dataset $\mathcal{D}_0$, and $i\in[0,1]$ corresponds to the $x$-axis of Fig.~\ref{fig:P1}(a). similarly,  Fig.~\ref{fig:P1}(c) shows the results of testing models  $\mathcal{W}_i$ on $\mathcal{D}_2$, and $i\in[2,10]$.  Fig.~\ref{fig:P1}(l) and Fig.~\ref{fig:P1}(m) shows the performance of testing $\mathcal{W}_i$ ($i\in[0,10]$) on the ROSE-YOUTU and CeFA datasets respectively. Fig.~\ref{fig:P2} is also organized in the same way to show the results of Protocol-2. Through experiments as shown in  Fig.~\ref{fig:P1} and Fig.~\ref{fig:P2}, we obtain the below observations. 

\vspace{-0.8em}
\noindent\textbf{Observation 1:} Significant catastrophic forgetting usually occurs when there is a significant domain gap between a new domain and previous domains. If a new domain has shared knowledge with previous domains, previous knowledge can be recalled. To analyze the significant catastrophic forgetting, we use ViT-Adapter in Fig.~\ref{fig:P1}(b), (c), and (d) as examples. In Protocol-1, $\mathcal{D}_1$ (REPLAY-ATTACK), $\mathcal{D}_2$ (CASIA-FASD), and $\mathcal{D}_3$ (MSU MFSD) are all 2D attack (Photo, Replay) datasets. In Session 4, the new coming domain dataset HKBU MarV2 ($\mathcal{D}_4$) only contains 3D Mask attack, which is significantly different from the 2D attack datasets. Therefore, if we observe the curve of vanilla ViT-Adapter in Fig.~\ref{fig:P1}(b), (c), and (d), after the training on $\mathcal{D}_4$(HKBU) in Session 4, the previous knowledge about REPLAY-ATTACK, CASIA-FASD and MSU-MFSD are significantly forgotten. Therefore, in Session 4 of Fig.~\ref{fig:P1}(b), (c), and (d), the AUC performance dramatically drops from Session 3. 
Besides forgetting, previous knowledge can be recalled when the new data domain has shared knowledge with the data in the previous domain. For example, in Fig.~\ref{fig:P1}(h), the new coming dataset $\mathcal{D_7}$ is the CASIA-SURF dataset \cite{CASIA-SURF}. After Sessions 8 and 9, AUC performance (black curve) obtained by ResNet18 drops significantly. In Session 10, the AUC performance experiences a downtrend first because of the forgetting issue, and then this performance increases to that of Session 8. By analyzing the datasets $\mathcal{D}_7$, $\mathcal{D}_8$, $\mathcal{D}_9$, and $\mathcal{D}_{10}$ in Protocol-1, we find that $\mathcal{D}_7$ (CASIA-SURF \cite{CASIA-SURF}) and $\mathcal{D}_{10}$ (CASIA-SURF 3D Mask) mainly contain Asian subjects. In contrast, the WFFD~\cite{WFFD} and WMCA~\cite{george2019biometric} mainly contain Caucasian subjects. Thus, there is less domain shift between CASIA-SURF and CASIA-SURF 3DMask. In Session 10, the model learns on CASIA-SURF 3DMask and recall the knowledge of CASIA-SURF. Similarly, in Protocol-2, $\mathcal{D}_1$, $\mathcal{D}_2$, $\mathcal{D}_3$, and $\mathcal{D}_4$ are the CASIA-SURF 3DMask, WMCA, WFFD and CASIA-SURF datasets respectively. As shown in Fig.~\ref{fig:P2}(b), the ResNet18 also forgets previous knowledge of $\mathcal{D}_1$ in Session 2 and 3, but recalls it in Session 4. Similar situations can be observed with other backbones.

\begin{table}[tbp]
\vspace{-0.4em}
  \centering
  \caption{Results of finetuning (FT) and our PPCR in optimizing models in our DCL-FAS setting. Better performance (\%) in comparison is in \textbf{bold}. }
  \vspace{-0.2em}
   \resizebox{1.0\linewidth}{!}{
    \begin{tabular}{clc|c|ccc|c|c}
    \hline
    \multirow{2}[4]{*}{Metric} & \multicolumn{1}{c}{\multirow{2}[4]{*}{Method}} & \multicolumn{3}{c}{Protocol-1} &       & \multicolumn{3}{c}{Protocol-2} \bigstrut\\
\cline{3-5}\cline{7-9}          &       & \multicolumn{1}{c}{ViT-ConvA} & \multicolumn{1}{c}{ViT-CDCA} & \multicolumn{1}{c}{ViT-DCDCA} &       & \multicolumn{1}{l}{ViT-ConvA} & \multicolumn{1}{c}{ViT-CDCA} & \multicolumn{1}{c}{ViT-DCDCA} \bigstrut\\
    \hline
    \multirow{2}[1]{*}{$mAA$ (\%)} & FT    & \textbf{93.29} & \textbf{93.06}& 93.23  &       & \textbf{94.65} & \textbf{94.42} & 93.79 \bigstrut[t]\\
          & PPCR (ours)  & 93.04& 92.12& \textbf{93.54} &       & 93.91 & 94.26  & \textbf{94.03} \\
          \hline
    \multirow{2}[1]{*}{$mABT$ (\%)} & FT    & -4.24& -4.08& -3.59  &       & -2.28  & -1.89  & -1.82 \\
          & PPCR (ours) & \textbf{-3.08}& \textbf{-3.92}& \textbf{-3.29}  &       & \textbf{-1.76}  & \textbf{-1.25}  & \textbf{-1.72} \bigstrut[b]\\
    \hline
    \multirow{2}[2]{*}{$mAGA$ (\%)} & FT    & 76.73 & 77.01 & 78.70  &       & 73.79 & 74.48 & 78.09 \bigstrut[t]\\
          & PPCR(ours)  & \textbf{80.86} & \textbf{77.17} & \textbf{82.06} &       & \textbf{79.14} & \textbf{75.73} & \textbf{80.08} \bigstrut[b]\\
    \hline
    \end{tabular}%
    }
    \vspace{-0.5cm}
  \label{tab:COMPARE PPCR-and-FT}%
\end{table}%

\vspace{0.3em}
\noindent\textbf{Observation 2:} \rz{A model that generalizes better to unseen domains usually forgets less previous domain knowledge.} From Fig.~\ref{fig:P1}(l) and Fig.~\ref{fig:P2}(l), we can see the ResNet18 has much lower generalization performance on the ROSE-YOUTU dataset than our ViT-CDCA and ViT-DCDCA. Also, Fig.~\ref{fig:P1}(a)-(k) and Fig.~\ref{fig:P2}(a)-(k) show that ResNet18's performance fluctuates more than our ViT-CDCA and ViT-DCDCA, indicating that ResNet18 often forgets previous knowledge in continual learning. By contrast, the performance curve of our ViT-CDCA and ViT-DCDCA are more stable, meaning our ViT-CDCA and ViT-DCDCA forget less than ResNet18. Therefore, we observe that a more generalized model could forget less in continual learning. The insight behind this observation is intuitive. Although different FAS datasets have domain gaps between each other, there could be some shared knowledge and information. A generalized model can extract generalized knowledge across different datasets. Thus, when learning on a new dataset, the model can learn related cues about previous datasets, and thus it suffers from less catastrophic forgetting. Therefore, developing a model generalized to the unseen domain is essential for DCL-FAS.

\subsection{Quantitative Analysis}
\label{sec:Analysis2}
\noindent\textbf{Impact of generalized architectures.}
In this section, we analyze quantitatively the performance of finetuning different architectures in Protocol-1 and Protocol-2. As shown in Table~\ref{tab:comparebackbone}, we can see that ResNet18 has the worst performance of $mAA$, $mABT$, and $mAGA$, indicating ResNet-18 is not as generalized as the other ViT architectures. Also, as ViT-Adapter merely uses linear layers as adapters, it lacks  FAS-specific inductive bias to extract generalized FAS features. Thus, ViT-Adapter also achieves poorer generalization performance of $mAGA$ than ViT-ConvA/CDCA/DCDCA and suffers more severe catastrophic forgetting with higher negative values of $mABT$. Meanwhile, we can see the effectiveness of our ViT-DCDCA as it achieves better $mABT$ and $mAGA$ than ViT-ConvA and ViT-CDCA. Thus, our proposed method of dynamically adapting central difference cues can achieve better generalization performance on unseen domains and less forgetting in the domain continual learning setting.

\vspace{0.3em}
\noindent\textbf{Efficacy of PPCR.} 
In Table~\ref{tab:COMPARE PPCR-and-FT}, we compare models using finetuning (FT) or the proposed PPCR in continual learning. With different architectures, our PPCR can achieve significantly better $mABT$ and $mAGA$ than FT. Thus, our PPCR can help to generalize more and forget less than FT in DCL-FAS. In Table~\ref{tab:remove pp}, `CR' means using $\mathcal{L}_{Con}$ in the optimization but does not include the proxy prototypes. The performance of `CR' and `PPCR' are comparable in terms of $mAGA$, and  PPCR forgets less with better $mABT$ than CR in both Protocol-1 and Protocol-2. Thus, the proposed proxy protocol can help reduce forgetting.

\begin{table}[tbp]
  \centering
  \caption{Results of our ViT-DCDCA with Proxy Prototype (PPCR) and without Proxy Prototype (CR).}
 \resizebox{1.0\linewidth}{!}{
    \begin{tabular}{l|ccccccc}
    \hline
    \multicolumn{1}{c|}{\multirow{2}[4]{*}{Algorithm}} & \multicolumn{3}{c}{Protoco-1} &       & \multicolumn{3}{c}{Prococol-2} \bigstrut\\
\cline{2-4}\cline{6-8}          & $mAA$ (\%) & $mABT$ (\%) & $mAGA$ (\%) &       & $mAA$ (\%) & $mABT$ (\%) & $mAGA$ (\%) \bigstrut\\
    \hline
    CR & \textbf{93.71} & -3.57 & 81.87 &       & \textbf{94.11} & \-1.77 & \textbf{80.31} \bigstrut[t]\\
    PPCR  & 93.54 & \textbf{-3.48} & \textbf{82.06} &       & 94.03 & \textbf{-1.72} & 80.08 \bigstrut[b]\\
    \hline
    \end{tabular}%
    }
  \label{tab:remove pp}%
\end{table}%

\begin{table}[tbp]
  \centering
  \caption{Results of our ViT-DCDCA with PPCR, EWC and LWF.}
 \resizebox{1.0\linewidth}{!}{
    \begin{tabular}{clccccccc}
    \hline
    \multicolumn{1}{c}{\multirow{2}[2]{*}{Method}} & \multicolumn{3}{c}{Protocol-1} &       & \multicolumn{3}{c}{Protocol-2} \bigstrut[t]\\
                 & \multicolumn{1}{c}{$mAA$ (\%)} & \multicolumn{1}{c}{$mABT$ (\%)} & \multicolumn{1}{l}{$mAGA$ (\%)} &       & \multicolumn{1}{c}{$mAA$ (\%)} & \multicolumn{1}{c}{$mABT$(\%)} & \multicolumn{1}{l}{$mAGA$ (\%)} \bigstrut[b]\\
    \hline
  EWC\cite{ewc}   & 92.35 & -0.163 & 78.41 &       & 92.32 & -0.52  & 78.43 \bigstrut[t]\\
           LWF\cite{lwf}   & 91.61 & -\textbf{0.97}  & 78.68 &       & 91.22 & \textbf{0.14} & 77.85 \\
           PPCR (Ours)  & \textbf{93.54} & -3.48  & \textbf{82.06} &       & \textbf{94.03} & -1.72  & \textbf{80.08} \bigstrut[b]\\
    \hline
    \end{tabular}%
    }
    \vspace{-1.5em}
  \label{tab:compare ewc lwf}%
\end{table}%


\vspace{0.3em}
\noindent\textbf{Comparison with existing continual learning methods.}
We implement two benchmark continual learning methods EWC \cite{ewc} and LWF \cite{lwf} to train our ViT-DCDCA and make comparisons to ViT-DCDCA with our PPCR. EWC and LWF achieve better $mABT$ than our PPCR, as they are designed specifically in forgetting less in continual learning. However, both EWC and LWF do not consider the generalization capability, and they achieve significantly less $mAA$ and $mAGA$ performance than our PPCR.

\begin{figure}
    \vspace{1.5em}
    \centering
    \includegraphics[width=\linewidth]{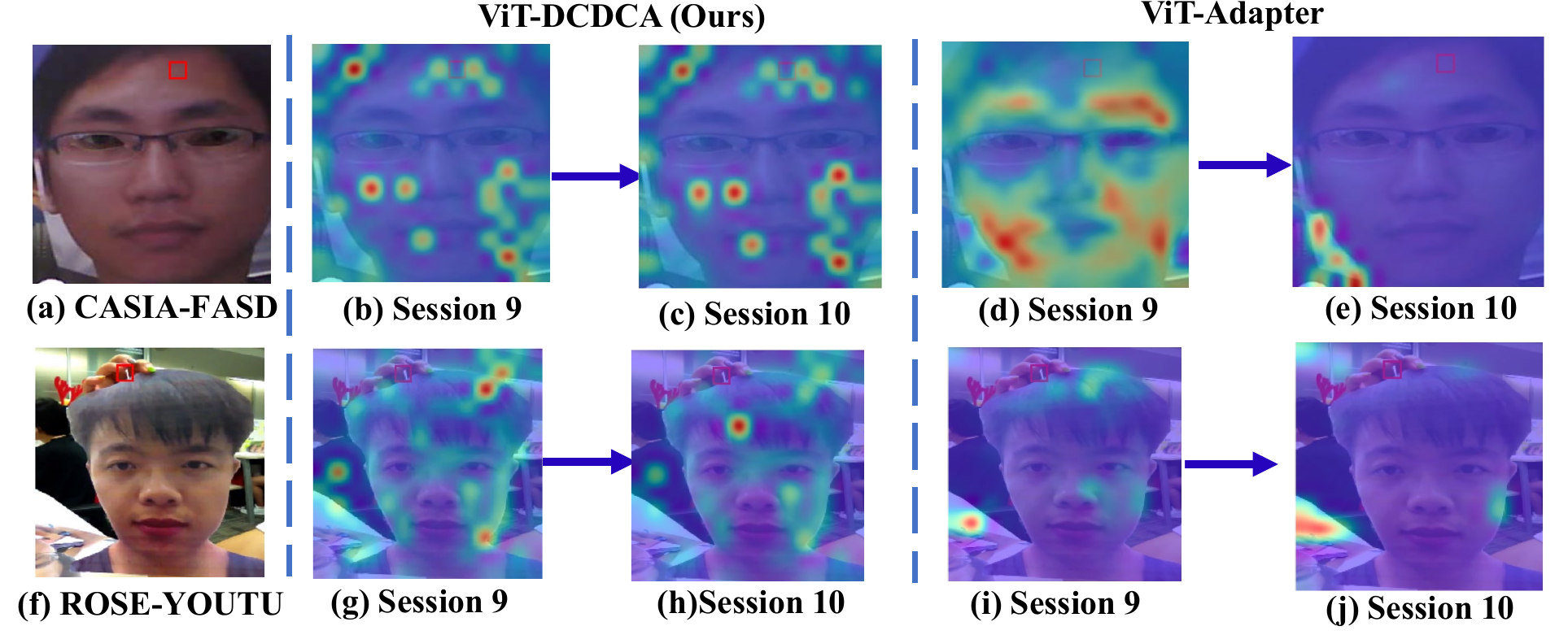}
    \vspace{-1.5em}
    \caption{Visual attention maps in Protocol-2. Red means high activation. (a) shows an attack example from $\mathcal{D}_9$ (CASIA-FASD) of Protocol-2. (b) and (c) are (a)'s attention maps from our ViT-DCDCA in Sessions 9 and 10, (d) and (e) are from ViT-Adapter. (f) is an attack example from the unseen ROSE-YOUTU. (g) and (h) are (f)'s attentions maps from our ViT-DCDCA in Sessions 9 and 10, and (i) and (j) are from ViT-Adapter.} \label{fig:vis}
     \vspace{-1.1em}
\end{figure}

\vspace{-0.4em}
\subsection{Visualization and Analysis}
\vspace{-0.5em}
In Fig.~\ref{fig:vis}, we visualize the attention maps to analyze the generalization and forgetting behavior of ViT-DCDCA and ViT-Adapter in Protocol-2. Face areas in Fig.~\ref{fig:vis}(b) and Fig.~\ref{fig:vis}(c) are highly activated. Moreover, the two attention maps are highly overlapped, meaning ViT-DCDCA's knowledge about CASIA-FASD is rarely forgotten from Session 9 to Session 10. By contrast, From Fig.~\ref{fig:vis}(d) to Fig.~\ref{fig:vis}(e), the activation maps changes dramatically, meaning the ViT-Adapter's knowledge about CASIA-FASD is largely forgotten, which corresponds to Fig.~\ref{fig:P2}(i) that ViT-Adapter forgets much more knowledge than our ViT-DCDCA about the CASIA-FASD dataset in Session 10. On the other hand, in Fig.~\ref{fig:vis}(g) and Fig.~\ref{fig:vis}(h), the paper edges areas are activated for spoofing classification by our ViT-DCDCA. Meanwhile, in Fig.~\ref{fig:vis}(i) and Fig.~\ref{fig:vis}(j), the background areas but not the face areas are activated, which indicates the corresponds to Fig.~\ref{fig:P2}(l) that ViT-Adapter achieves much worse generalization performance than ViT-DCDCA.

\vspace{-0.6em}
\section{Conclusion}
\vspace{-0.4em}
In this paper, we raise the concern of privacy in continual FAS and formulate the FAS in rehearsal-free Domain Continual Learning settings. \rz{We devise more practical protocols for evaluation and experimentally find that models with better unseen domain generalization can also have less forgetting during continual learning. For better generalization performance, we develop a Dynamic Central Difference Convolutional Adapter to adapt ViT continually. Besides, we propose Proxy Prototype Contrastive Regularization to provide previous knowledge from previous model weights to reduce forgetting. Extensive experiments on two protocols demonstrate that the proposed method generalizes better in the unseen domain and forgets less previous knowledge compared to baseline methods.}


{\small
\bibliographystyle{ieee_fullname}
\bibliography{egbib}
}

\end{document}